# E2F-Net: Eyes-to-Face Inpainting via StyleGAN Latent Space

Ahmad Hassanpour, Fatemeh Jamalbafrani, Bian Yang, Kiran Raja, Raymond Veldhuis, Julian Fierrez

Ahmad Hassanpour, Bian Yang, Kiran Raja, and Raymond Veldhuis are with Department of Information Security and Communication Technology NTNU, Gjovik, Norway (ahmad.hassanpour@ntnu.no, bian.yang@ntnu.no, kiran.raja@ntnu.no, raymond.veldhuis@ntnu.no)

F. Jamalbafrani is with Department of Electrical Engineering, Sharif University of Technology, Tehran, Iran (fatemeh_jamalbafrani@ee.sharif.edu)

J. Fierrez is with school of Engineering, Universidad Autonoma de Madrid, Madrid, Spain (julian.fierrez@uam.es)

**Abstract**

Face inpainting, the technique of restoring missing or damaged regions in facial images, is pivotal for applications like face recognition in occluded scenarios and image analysis with poor-quality captures. This process not only needs to produce realistic visuals but also preserve individual identity characteristics. The aim of this paper is to inpaint a face given periocular region (eyes-to-face) through a proposed new Generative Adversarial Network (GAN)-based model called Eyes-to-Face Network (E2F-Net). The proposed approach extracts identity and non-identity features from the periocular region using two dedicated encoders have been used. The extracted features are then mapped to the latent space of a pre-trained StyleGAN generator to benefit from its state-of-the-art performance and its rich, diverse and expressive latent space without any additional training. We further improve the StyleGAN's output to find the optimal code in the latent space using a new optimization for GAN inversion technique. Our E2F-Net requires a minimum training process reducing the computational complexity as a secondary benefit. Through extensive experiments, we show that our method successfully reconstructs the whole face with high quality, surpassing current techniques, despite significantly less training and supervision efforts. We have generated seven eyes-to-face datasets based on well-known public face datasets for training and verifying our proposed methods. The code and datasets are publicly available[1].

*Keywords*: Eyes-to-Face, Face Inpainting, Face Reconstruction, GAN Latent Space, StyleGAN.

## 1. INTRODUCTION

Face inpainting is the process of approximating the missing or masked face elements using the auxiliary data from around of the missing region. Thus, estimating those missing regions is vital in practice, particularly in face recognition under occlusions, and in general any image/video analysis application on low quality, uncontrolled, or in-the-wild acquisition conditions. Realistic approximation or inpainting despite being highly applicable, is known to be particularly hard task. This is mainly due to high photometric, geometric and kinematic complexities, and because the human face contains numerous independent, high dimensional characteristics that are not easy to approximate and also make it realistic for human perception [1]. Like other image inpainting tasks (e.g., scenes inpainting [3, 4], streets inpainting [5, 12]), some key requirements for face inpainting are:

This work was supported by the Project Privacy Matters (PRIMA) under Grant H2020-MSCA-ITN-2019-860315 and Project BBforTAI under Grant PID2021-127641OB-I00 MICINN/FEDER. *(Corresponding author: A. Hassanpour).*

[1] https://github.com/fatemejamalii/E2F-Net

- **R1**) the filled region in corrupted area should be semantically meaningful in relation to the face,

- **R2**) the original content (unmasked) and approximated content should be continuously assembled and consistent,

- **R3**) the inpainted image should be visually realistic and have high fidelity.

Reconstructing the corrupted/unavailable portions of a face such that the topological consistency between facial attributes are preserved (both identity and non-identity[2] attributes), is not a trivial task [6, 8]. One can however exploit that human faces share common geometrical and appearance distributions, which are then personalized for given subjects in specific conditions. General face geometry/appearance models have been used to ease face manipulation and completion for given subjects [2]. Notably, a specific facial representation deviating from or sampling the general model can be considered for the purpose of identity information completion due to the unique topology of different facial elements and their distinctive characteristics.

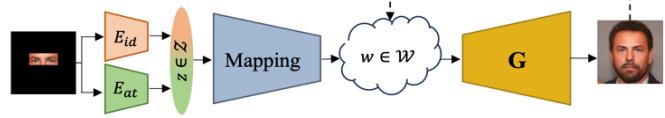

**Fig. 1.** The proposed face reconstruction framework utilizes two encoders called identity ($E_{id}$) and attribute ($E_{at}$), to generate the latent code $z$. The latent code $z$ is then mapped to the latent space $\mathcal{W}$ of a pre-trained generator shown by $G$. Finally, the output of $G$ will be refined by finding the optimal point in $\mathcal{W}$ space using an optimizer (return arrow from $G$'s output to $\mathcal{W}$ space).

Among all facial elements, the eyes are one of the most expressive organs on the human face and contain discriminative features [39]. In this paper, we aim to reconstruct a face using the periocular region alone which we refer to as eyes-to-face inpainting (see Fig. 1). Therefore, in addition to the above-mentioned challenges (i.e., **R1-R3),** another set of criteria for eyes-to-face inpainting are as follows:

- **R4**) The topological structure of the face should be reconstructed in such a way that all elements are placed in the proper position both semantically and continuously. For doing this, it is essential to predict the shape of the face precisely and place the face's elements (e.g., nose, chin, mouth) proportionally within the predicted frame. Moreover, the head pose should be aligned and integrated with other elements based on the appearance of eye.

- **R5**) The usefulness of skin around the eyes for determining demographic features (e.g., age, gender) has been shown in previous research [43]. The proposed inpainting model should therefore be able to estimate demographic characteristics using the color, texture, and size of the eyes and brows, then inpaint other facial attributes using the predicted attributes.

- **R6**) The proposed solution should preserve the identity-related features present in the eyes region when reconstructing the whole

---

[2] Here, identity features are facial features that can be used to verify the identity of a person using his face including demographic characteristics (e.g., age, gender, color skin), and non-identity features are characteristics like head pose, and facial expression.

face.

It is also important to note that the performance of suggested solutions can be directly impacted by the image's masking, and it is obvious that bigger masks make it harder to achieve the referenced requirements (i.e., **R1**–**R6**).

Generally, synthetic and natural masks are considered in face inpainting scenarios, which can be classified into two categories called free- (irregular) and fixed-form (regular) masks. In widely used free-form masks [23, 24], there are irregular shapes randomly placed on the images (Fig. 2(a)(b)), useful for inpainting irregular scratches. Instead, in the fixed-form masks, regular shapes cover some portions of the images which are placed on the images randomly or purposefully (Fig. 2(c)-(e)) [21, 22, 25].

Recently, learning-based techniques such as deep convolutional neural networks (CNNs) and generative adversarial networks (GANs) have been widely used for a variety of image inpainting tasks, such as eliminating objects [7, 8], noises [9], texts [10], and masks [11]. Usually, the proposed CNN-based methods are classified into three categories including coarse-to-fine, coarse-and-fine, and structural guidance-based methods. Coarse-to-fine based methods [13, 14] exploit two-stage architectures to complete content formation and texture refinement. A two-stage design produces an intermediate coarse image after reconstructing structures in the first step, then feeds it to the second stage for texture improvement. The second category called coarse-and-fine [15, 16] consists of two parallel branches, i.e., coarse and fine, that extract coarse and fine information simultaneously and fill the missed regions using the extracted information. The final group of approaches, known as structural guidance-based methods, employs an assistance algorithm to provide additional information, such as edges [17, 18], contours [19], or landmarks [6], for the proposed inpainting method.

The rest of the paper is organized as follows. Recent works in face inpainting, latent space embedding, and GAN inversion are reviewed in Section 2. Limitations of related works and our contributions have been discussed in section 3. A detained description of the proposed method is provided in Section 4. The experimental results and ablation study are reported in Sections 5 and 6. Finally, sections 7 and 8 present discussion and conclusion.

## 2. Background and Related Work

In this section, we briefly review the most relevant research on face inpainting, latent space embedding, and GAN inversion in the following subsections.

### 2.1. Face Inpainting

There are a few works that particularly attempted to reconstruct a face using the periocular region. Luo et al. [27] proposed a three-step solution called EyesGAN which includes two GANs to predict other parts of a face using the eyes region. They proposed a self-attention mechanism to extract informative and attention feature maps from convolution layers. Unfortunately, it is difficult to compare our results to EyesGAN due to its unavailability. Hassanpour et al. [18] proposed a GAN-based coarse-to-fine method called E2F-GAN such that the coarse module benefits from the coarse-and-fine architecture. They used edges as the guidance information for the designed coarse-to-fine network.

With the aim of face inpainting by placing randomly regular or irregular masks, several methods have been proposed recently. In order to produce more realistic images, Chen et al. [28] presented a generative-based coarse-to-fine structure that takes advantage of an attention layer to capture lengthy dependencies between features. Free-form masks are inpainted using a coarse-to-fine structure proposed by Yu et al. [13]. A novel attention layer in a coarse-to-fine design was suggested by Liu et al. [30] in the same context. Wang et al. [24] proposed a two-stage face inpainting method to detect the corrupted regions and then improve inpainting results using a top-down refinement network.

A few works proposed guidance-based techniques. An edge generator is used by Nazari et al. [17] to reconstruct the edges before feeding the corrupted image and predicted edges to the image inpainting network. In order to extract features and recover the structures and textures of missing regions, Chen and Liu [19] employ a dual-branch network with texture and edge branches. Some works estimate facial landmarks to assist the main inpainting network[6]. A unique output per each input is generated by the methods indicated above. In contrast, some other approaches inpaint the corrupted regions differently for each specific input. A dual pipeline based on Variational Auto-Encoders (VAEs) was proposed by Zheng et al. [26], with a reconstructive path that uses the ground truth to learn the prior distribution of missing regions and a generative path for which the conditional prior is connected to the distribution learned in the reconstructive path. Zhao et al. [32] have suggested a GAN-based unsupervised conditional framework for different image inpainting that can learn conditional completion distributions.

*2.2. Latent Space Embedding*

With the rapid evolution of GANs, many works have tried to understand and control their latent space for various image editing tasks [33]. Choosing which latent space to embed an image into a GAN image generator is a crucial design decision for editing flexibility and output quality. One of the most successful approaches for generating this embedding was described in the framework of StyleGAN, which has been followed extensively in the recent past [34, 35]. By using an 8-layer multilayer perceptron (MLP) to create a nonlinear mapping network $M$, StyleGAN [36] transforms a native $z \in \mathcal{Z}$ to a style vector $w \in \mathcal{W}$. The $\mathcal{W}$ space is the name given to this intermediate latent space. The $\mathcal{W}$ space of StyleGAN contains more disentangled characteristics than the $\mathcal{Z}$ space does because of the mapping network $M$.

*2.3. GAN Inversion*

GAN inversion tries to invert a given image back into a pretrained GAN model's latent space. The generator can then accurately rebuild the image from the inverted code. Learning-based, optimization-based, and hybrid methods are the three major strategies for GAN inversion with the purpose of projecting images into the latent space. Learning-based GAN inversion [39] typically involves training an encoding neural network $E(x; \theta_E)$ to map an input image, $x$, into the latent code $z$:

$$\theta_E^* = \arg_{\theta_E} \min \sum_n \mathcal{L}\big(G\big(E(x_n; \theta_E)\big), x_n\big) \tag{1}$$

where $x_n$ denotes the $n$-th image in the training dataset and $z = E(x; \theta_E^*)$. The objective in (1) is reminiscent of an autoencoder pipeline, with an encoder $E$ and a decoder $G$. Throughout the training, the decoder $G$ remains fixed. Furthermore, improving the latent vector is generally used to reconstruct a target image by optimization-based GAN inversion approaches [40].

$$\mathbf{z}^* = \arg_z \min \mathcal{L}(x, G(z; \theta_G)) \tag{2}$$

where $x$ is a target image and $G$ is a GAN generator parameterized by $\theta_G$. The hybrid methods [41] exploit the advantages of both previously described approaches adjusting both the Encoder $\theta_E^*$ and the specific location in the latent space $z^*$.

*3. Limitations of Related Works and Our Contributions*

The existing face inpainting works use different strategies (e.g., coarse-to-fine, coarse-and-fine, guidance information) to address **R1-R3**. The coarse-to-fine structure has two limitations. First, the coarse result has to be reasonably accurate for an effective refinement, and second, the cascaded dilated convolutions smooth the details of features, resulting in blurry inpainting results [28]. Although structural information about the target image may assist and increase the performance of the inpainting generator in some cases, estimating that information can also slow the inference speed, increase the computational cost, and introduce the necessity of handcrafting auxiliary information (e.g., edge, contour, or landmarks) for different applications when guidance-based methods are used. Moreover, unlike approaches used in [26, 32], an eyes-to-face approach should generate a unique output for each input even after several executions to fulfill the requirements **R5** and **R6**.

We address these requirements and limitations by proposing a novel method to reconstruct a face using features extracted from the periocular region. A major part of inpainting is solely done using relevant pre-trained networks eliminating the need for additional training. The overall architecture of our proposed framework (see Fig. 1) resembles coarse-and-fine architecture, but differs in several ways, detailed in the following. Our key idea is to directly map the extracted latent representation to the latent space of a pre-trained generator, as depicted in Fig. 1. To extract a latent representation which includes identity (ID) and non-identity (non-ID) attributes from the periocular region, we use a pre-trained face recognition method, shown by $E_{id}$, and a trainable network, shown by $E_{at}$ respectively. We then map the resulting latent code $\mathcal{Z}$ to the latent space $\mathcal{W}$ of a pre-trained generator $G$, and evaluate the quality of the inpainting only on the $G$'s output. specifically, we use $\mathcal{W}$ space to convert the extracted ID and non-ID features into more disentangled space, and the $\mathcal{Z}$ space is created by concatenating ID and non-ID characteristics instead of using a Gaussian distribution. An optimizer is further used to find the optimal point in the $\mathcal{W}$ space based on the output of $G$ in the last step as optimization-based GAN inversion technique, leading to address **R4-R6** more precisely. This mapping empowers us to utilize a state-of-the-art pre-trained generator, inheriting its high-resolution and output diversity, with minimum training process. In our approach, the representation is split into two segments comprising separate and meaningful information (i.e., ID and non-ID information). Then the mapping network ($M$) is trained to extract the relevant information from the output of $E_{id}$ and $E_{at}$ to be merged into a proper representation of the target face. We will show that our method can effectively perform this task and inpaint the hidden region with high quality.

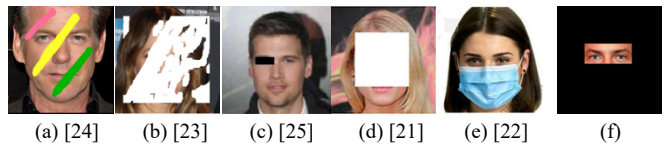

(a) [24]  (b) [23]  (c) [25]  (d) [21]  (e) [22]  (f)

**Fig. 2.** Examples of two types of widely used masks: free-form masks a (b), and fixed-form (c)-(e). The mask used in this work is shown in (f).

Further, we employ a large-size mask that covers about 75% of the face image since the goal of this paper is to complete the face based on the region of the eyes, unlike other existing works. Despite using large-size masks, we do not use any guidance information during our training in our work as compared to other related works [6, 17, 18, 19] to reduce training time and increase inference speed. Like other face inpainting methods, the performance of our face inpainting is dependent on the capabilities of the selected generator. Using StyleGAN as generator, the output of our proposed methods benefits from high image quality, outperforming all previous face inpainting methods we have compared against. In addition to being of the highest quality, our technique also successfully generates the entire face with realistic hair region, which is reported to help in identification tasks. Therefore, in contrast to state-of- the-art face inpainting methods, which need to train one or more generators [18, 26], we use a pre-trained generator reducing the training efforts. To validate our proposed method, several qualitative and quantitative metrics have been evaluated and compared with four state-of-the-art methods. Our experiments not only assess the quality of inpainted regions but also estimated demographic and ID features. Moreover, the effectiveness of reconstructing and maintaining identification elements on unseen faces, as well as the quality and diversity of faces, have been compared across all methods. Our approach has been demonstrated to outperform earlier work in addition to providing special benefits including the reconstructing of the full head and hair, preservation of ID and non-ID traits, and minimum supervision, which eliminates the need for a substantial training set.

It should be noted that different face elements impacts face recognition to varying degree as assessed in several works [54]. In this work, we attempt to extract the ID information that existed in the periocular region and preserve it in the reconstructed image. Our results show face recognition performance using inpainted images provides better accuracy than the periocular region alone, indicating our proposed algorithm not only preserves the ID information from the periocular region but also it can predict the dependent ID information and add it to reconstructed face for further recognition tasks. Further, seven new datasets of masked faces called E2F-StyleGANdb, E2F-CelebA-HQ, E2F-FFHQ, E2F-MS1MV2, E2F-LFW, E2F-CFP-FP, E2F-AgeDB-30 have been generated to train and evaluate our proposed method. Additionally, to measure the ID information in the reconstructed image, we generated two other datasets, described in section 4.

Table 1 presents a comparative analysis of our methodology against other related studies, focusing on key attributes essential for applications involving eye-to-face reconstruction.

**Table 1.** Comparative analysis of face inpainting methods.

| Reference | one-to-one mapping | auxiliary independence | bio-facial reconstruction | high-resolution output (1024 × 1024) | pre-trained generator | focused on eyes-to-face | fulfilled requirements | mask type (coverage ratio) | Used losses |
|---|---|---|---|---|---|---|---|---|---|
| [17] | ✓ | × | × | × | × | × | **R1- R4** | free-form (30-60%) | Perceptual Loss, Hinge loss, $L_1$ Loss, Style Loss, Adversarial Loss |
| [6] | ✓ | × | × | × | × | × | **R1- R4** | free-form (30-60%), fixed form (50%) | Perceptual Loss, $L_2$ Loss, Style Loss, Total variation loss, Adversarial Loss |
| [28] | ✓ | ✓ | × | × | × | × | **R1- R4** | fixed form (50%) | Hinge loss, Adversarial Loss |
| [13] | ✓ | ✓ | × | × | × | × | **R1- R4** | fixed form (50%) | $L_1$ Loss, Reconstruction Loss, Adversarial Loss |
| [19] | ✓ | × | × | × | × | × | **R1- R4** | free-form (30-60%) | $L_1$ Loss, Adversarial loss |
| [32] | × | ✓ | × | × | × | × | **R1- R4** | fixed form (50%) | KL loss, Reconstruction Loss, Adversarial Loss |
| [26] | × | ✓ | × | × | × | × | **R1- R4** | free-form (30-60%), fixed form (50%) | $L_2$ Loss, KL loss, Adversarial Loss |
| [27] | ✓ | × | × | × | × | ✓ | **R1- R6** | fixed-form (75%) | Perceptual Loss, $L_1$ Loss, $L_2$ Loss, KL loss, Adversarial Loss |
| [18] | ✓ | × | ✓ | × | × | ✓ | **R1-R6** | fixed-form (75%) | Perceptual Loss, Style Loss, Reconstruction Loss, Adversarial Loss |
| ours | ✓ | ✓ | ✓ | ✓ | ✓ | ✓ | **R1- R6** | fixed-form (75%) | Perceptual Loss, Style Loss, Identity Loss, Landmark Loss, Reconstruction Loss, Adversarial Loss |

## 4. Proposed Method

### 4.1. Overview

As shown in Fig. 3, given a ground-truth face image $\mathbf{I}_{gt} \in \mathbb{R}^{h \times w \times 3}$, and a binary mask $\mathbf{I}_m \in \mathbb{R}^{h \times w \times 1}$ (with value 1 for known pixels and 0 for unknown pixels), the input image $\mathbf{I}_{in} \in \mathbb{R}^{h \times w \times 3}$ is obtained as $\mathbf{I}_{in} = \mathbf{I}_{gt} \odot \mathbf{I}_m$, where $\odot$ denotes the Hadamard product. The goal is to inpaint the whole face with preserving ID and other visual attributes, specifically pose, expression, and properly placing face elements with proper size. To extract ID and other attributes, we used two encoders denoted as $E_{id}$ and $E_{at}$ whose outputs are concatenated into $z$ (i.e., $z = [E_{id}(\mathbf{I}_c), E_{at}(\mathbf{I}_{in})]$). Then we map the $\mathcal{Z}$ space to a new space called $\mathcal{W}$, and the new representation $w$ feeds a generator. The generator generates a face based on both the ID and other facial attributes. Finally, we use an optimizer to ensure that the optimal point has been chosen in the $\mathcal{W}$ space to be fed to the generator. As depicted in Fig. 3, the proposed E2F-Net consists of two encoders $E_{id}$ and $E_{at}$, a mapping network $M$, a generator network called $G$ (StyleGAN). A few additional pre-trained encoders are used for calculating corresponding multiple losses as described afterwards: feature encoder ($E_{feat}$), landmark encoder ($E_{lnd}$), and face encoder ($E_{face}$).

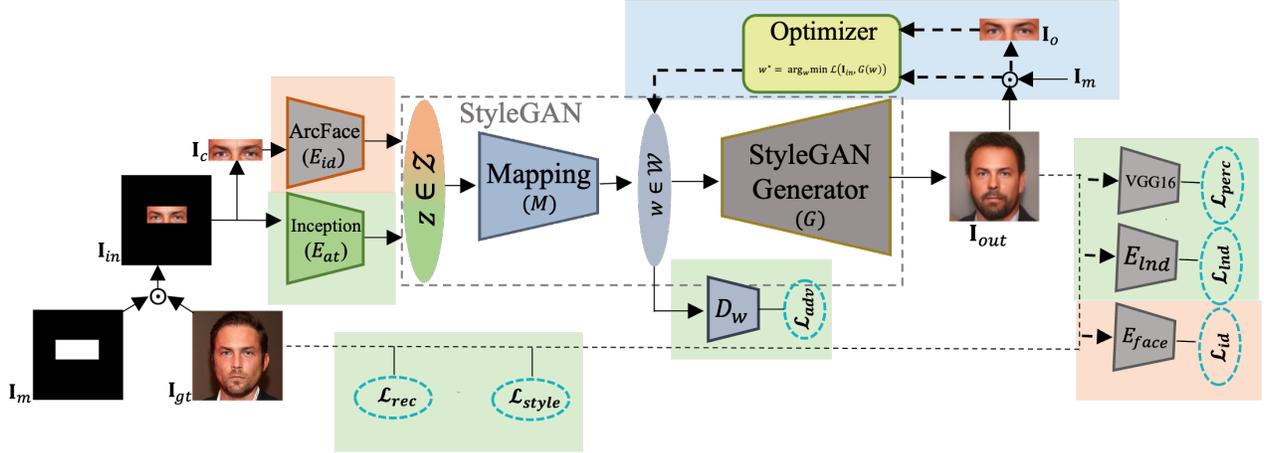

**Fig. 3.** The overview of our proposed reconstructing face method (E2F-Net). Data flow and losses show by solid lines and dashed ones, respectively. First, the ID and non-ID features are extracted from masked ($\mathbf{I}_{in}$) and cropped-masked ($\mathbf{I}_c$) images using encoders $E_{id}$ and $E_{at}$, respectively. Through our mapping network $M$, the concatenated features are mapped to $\mathcal{W}$, the latent space of the pre-trained StyleGAN generator $G$. Finally, the optimal latent code in $\mathcal{W}$ space has been found using an optimizer. The blue, orange, and green highlights indicate our major contributions. The **R1**–**R4** are being addressed by the green blocks. The block highlighted with blue is addressing **R4**–**R6**. The orange modules emphasize on **R5** and **R6**.

Notably, we use a state-of-the-art high-quality synthesize face generator called StyleGAN as the pretrained generator for all our experiments. Different from other GANs, StyleGAN features two latent spaces: $\mathcal{W}$, which is induced by a learned mapping from $\mathcal{Z}$, and $\mathcal{Z}$, which is generated by a fixed distribution. Since $\mathcal{W}$ is a more disentangled latent space than $\mathcal{Z}$ and is more suited to facilitate and accommodate image inpainting, we employ it to map the combined face code into it. We reduce the difficulty of learning to produce high-quality and high-fidelity images by employing this cutting-edge generator ($G$). However, it is not simple to train the mapping between the latent space of the encoders ($\mathcal{Z}$) and $\mathcal{W}$. To assist $M$ in anticipating features that lie within $\mathcal{W}$, we add a discriminator ($D_w$). To distinguish between real samples from StyleGAN's $\mathcal{W}$ space and $M$'s predictions, $D_w$ is trained in an adversarial manner.

### 4.2. The Architecture of the Proposed Method

The proposed E2F-Net has only three trainable modules: $E_{at}$, $M$, and $D_w$. The $E_{id}$ encoder is a pre-trained face recognition model called ArcFace, trained on the edited version of the MS1MV2 dataset called E2F-MS1MV2, described in the next subsection. The $E_{at}$ encoder is implemented as InceptionV3 [49]. The $M$ and $D_w$ both include four fully connected layers.

The generator, $G$ is a pre-trained StyleGAN synthesis network, trained on FFHQ [36]. In the following subsections, we will explain each used module in detail.

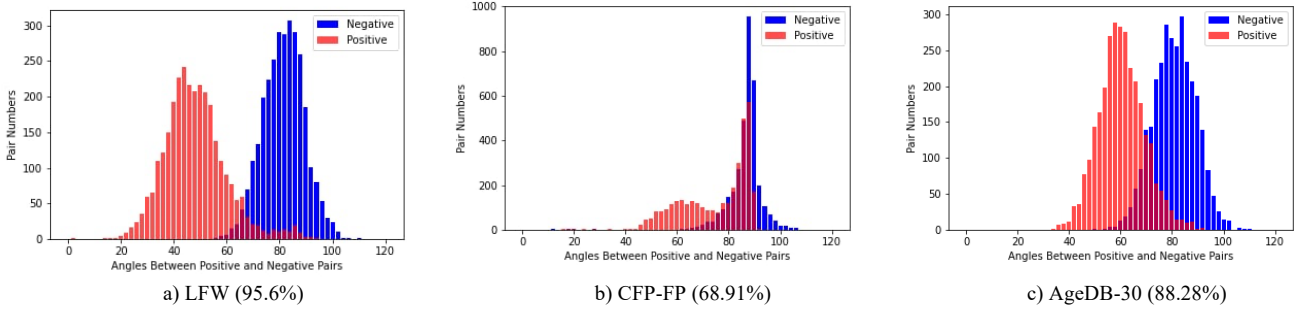

**Fig. 4.** Angle distributions of both positive and negative pairs on LFW, CFP-FP, and AgeDB-30. Red area indicates positive pairs while blue indicates negative pairs. All angles are represented in degrees.

4.2.1. *Identity Encoder*: To extract ID features from the periocular region, we utilized a face recognition model called ArcFace with a Resnet-50 backbone. To ensure that the features provided by ArcFace are well adapted to the periocular region, we retrained ArcFace model on a modified version of the MS1MV2 dataset called E2F-MS1MV2. To generate this dataset, all images in the MS1MV2 dataset were cropped to keep only the periocular region. By doing this, the recognition task is enforced to used eyes region alone. To validate the effectiveness of this model, we illustrate the angle distributions of both positive and negative pairs on edited versions of LFW, CFP-FP, and AgeDB-30, called E2F-LFW, E2F-CFP-FP, and E2F-AgeDB-30, in Fig. 4. We can see that the periocular region can be very effective for face verification task, with verification accuracies for the three datasets E2F-LFW, E2F-CFP-FP, and E2F-AgeDB-30 resulting in 95.6%, 68.91%, and 88.28%, respectively. Given the trained network parameters ($\theta_{id}$), the attribute encoder ($E_{id}$) is fixed and used to obtain the attribute code $z_{id} \in \mathbb{R}^{512 \times 1}$, i.e., $z_{id} = E_{id}(\mathbf{I}_c; \theta_{id})$.

4.2.2. *Attribute Encoder:* $E_{at}$ extracts non-ID features like pose, expression, illumination, skin color, etc. We used a pre-trained version of InceptionV3 [49] which has been trained on a large classification image dataset called ILSVRC 2012. Given the pre-trained network parameters ($\hat{\theta}_{at}$), the attribute encoder ($E_{at}$) is used to obtain the attribute code $z_{at} \in \mathbb{R}^{2048 \times 1}$, i.e., $z_{at} = E_{at}(\mathbf{I}_{in}; \theta_{at} | \hat{\theta}_{at})$. From the pre-trained parameters $\hat{\theta}_{at}$ we finetune until the final $\theta_{at}$ using appropriate loss functions as described in the next section.

4.2.3. *Mapping Network:* A multi-layer fully-connected neural network $M$, linearly maps the concatenated ID and non-ID attribute latent codes i.e., $z_{id}$ and $z_{at}$, $z \in \mathbb{R}^{2560 \times 1}$, to a stochastic style code $w \in \mathbb{R}^{512 \times 1}$, where $w$ lies in an extended stochastic latent space ($\mathcal{W}$). Let $\theta_M$ be the learnable parameters in $M$, then we have $w = M(z; \theta_f)$. Notably, $M$ is comprised of four fully connected layers. Other network sizes are explored in the ablation section.

4.2.4. *StyleGAN:* In addition to producing impressively photorealistic, high-quality synthetic photos of faces, StyleGAN, an extension to the GAN architecture, proposes significant changes to the generator model and allows to control over the style of the generated image at various levels of detail by adjusting the style vectors and various noise parameters. Given the pre-trained StyleGAN network parameters ($\theta_G$), the reconstructed face ($\mathbf{I}_{out}$) is the output of $G$ i.e., $\mathbf{I}_{out} = G(w; \theta_G)$, with $\mathbf{I}_{out} \in \mathbb{R}^{h \times w \times 3}$.

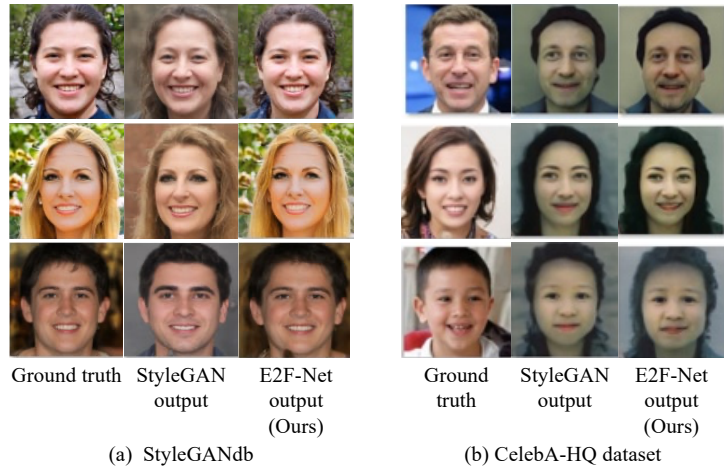

(a) StyleGANdb          (b) CelebA-HQ dataset

**Fig. 5.** The middle column shows the output of StyleGAN before optimization and the right column after the optimization proposed in our E2F-Net.

*4.2.5. Discriminator:* The introduction of StyleGAN as a module in our method provides significant benefits (e.g., high realism of the output), but also comes with some challenges. In particular, it is not simple to train the mapping between the latent space $\mathcal{Z}$ and $\mathcal{W}$. To help $M$ estimate features that lie within $\mathcal{W}$, we add a discriminator $D_w$, which is trained in an adversarial manner to distinguish between real samples from StyleGAN's $\mathcal{W}$ space and $M$'s predictions. Note that we used E2F-StyleGANdb dataset for training $D$ since we have latent code $w$ for each sample and during training with E2F-CelebA-HQ we do not train this module.

*4.2.6. Inversion via Optimization:* To fully exploit the ability and explore the interpretability of well-trained StyleGAN models, GAN inversion has been proposed to find the optimal latent codes within $\mathcal{W}$ space. In the optimization section, we generate the reconstructed face by optimizing over the latent vector $w$:

$$w^* = \arg_w \min \mathcal{L}\big(\mathbf{I}_{in}, G(w)\big) \tag{3}$$

where $\mathbf{I}_{in}$ is target image and $G(w)$ is the output of StyleGAN generator. Equation (3) is a non-convex optimization problem. The used loss functions for finding the optimum $w$ have been defined in the next section.

| **Algorithm 1**: Latent Space Embedding for StyleGAN |
| --- |
|    **Input**: $\mathbf{I}_{out}$ StyleGAN output, $\mathbf{I}_{in}$ masked input, $\mathbf{I}_c$ cropped input; $G$ the pre-trained StyleGAN generator . |
|    **Output**: optimum latent code $w^*$ |
| 1   Initialize latent code $w^* = w$ |
| 2   **while** *not converged* **do** |
| 3       $\mathbf{I}_{out_c} \leftarrow$ **Cropped** $G(w^*)$ |
| 4       $\mathbf{I}_{out_m} \leftarrow$ **Masked** $G(w^*)$ |
| 5       $\mathcal{L}_{opt} \leftarrow \mathcal{L}_{perc}(\mathbf{I}_{in}, \mathbf{I}_{out_m}) + \mathcal{L}_{id}(\mathbf{I}_c, \mathbf{I}_{out_c})$ |
| 6       $w^* \leftarrow w^* - \eta \ \nabla_w \mathcal{L}_{opt}$ |
| 7   **end** |

### 4.3. Training and Losses

First, a synthetic dataset using StyleGAN called E2F-StyleGANdb has been created in the following manner. We sample 50,000 random Gaussian vectors and forward them through the pre-trained StyleGAN. Then the periocular region has been cropped for each image generated from the vectors. The Gaussian noise is transformed into a latent vector $w$ in the forward process, from which we crop the image and capture both the image and the $w$ vector. The E2F-StyleGANdb images are split into two parts for training and verifying (90% and 10% respectively) the proposed model. During the training, the latent vectors $w$ are used as "real" samples for training the trainable modules. Fig. 5 (a) shows the generated results by E2F-Net which are very close to the ground truth. Despite accurate results for the E2F-StyleGANdb dataset, this behaviour is not seen presented in real-world scenarios. To examine this, a modified version of the CelebA-HQ dataset [48] called E2F-CelebA-HQ has been created. As shown in Fig. 5 (b), the gender and age of the person are preserved but the quality of outputs and identity of the person have not been preserved very well. To overcome this, the E2F-CelebA-HQ dataset has been used for training. The latent vectors have been obtained for all training samples of E2F-CelebA-HQ dataset by passing through $E_{id}$, $E_{at}$, and $M$. Similar to the previous attempt, the latent vectors $w$ are used as "real" samples for training the trainable modules.

It is noteworthy to note that the E2F-Net model is trained in a supervised, end-to-end fashion. To achieve noteworthy results, we have used a variety of loss functions, including identity loss, landmark loss, perceptual loss, style loss, adversarial loss, and reconstruction loss for trainable components of our proposed method. More specifically, the adversarial loss $\mathcal{L}_{adv}$ ensures proper mapping to the $\mathcal{W}$ space. Identity preservation is encouraged using $\mathcal{L}_{id}$, that penalizes differences in identity between $\mathbf{I}_{gt}$ and $\mathbf{I}_{out}$. Attributes preservation is encouraged using $\mathcal{L}_{rec}$ and $\mathcal{L}_{lnd}$, which penalize pixel-level and facial landmarks differences, respectively, between $\mathbf{I}_{gt}$ and $\mathbf{I}_{out}$. In the following we describe all losses.

*Perceptual Loss:* Perceptual loss [51] has been utilized to guarantee the similarity of high-level structures to keep the structure information of the overall image. Therefore, instead of matching pixels between them, similar feature representations to the ground truth are required to achieve **R1-R4**. We calculate the perceptual loss ($\mathcal{L}_{perc}$) by feeding the generated image ($\mathbf{I}_{out}$) and the ground truth ($\mathbf{I}_{gt}$) in a VGG-19 feature extractor. We then obtain feature maps $\varphi^{gt}$ and $\varphi^{out}$, extracted from layer $l$ of the pre-trained VGG19 network. The perceptual loss can be written as follows:

$$\mathcal{L}_{perc} = \sum_{l=1}^{N} \frac{\left\| \varphi_l^{gt} - \varphi_l^{out} \right\|_1}{C_l H_l W_l} \tag{4}$$

where $H_l$, $W_l$, and $C_l$ represent the height, weight, and channel size of the $l^{th}$ feature map, respectively. $N$ is the number of feature maps that the VGG-19 feature extractor generates.

*Style Loss*: Perceptual loss aids in obtaining high-level structure and prevents the output image from deviating in content from the ground truth which assist to enhancing **R1-R4**. We still need to maintain consistent style elements like colors and patterns, though. This objective can be achieved by adding style loss ($\mathcal{L}_{style}$) to the loss function. Similar to $\mathcal{L}_{perc}$, $\varphi^{gt}$ and $\varphi^{out}$ are extracted from VGG-19, and we define $\varphi_l^{style}$ as the product of a features map (row vector) multiplied by its transpose:

$$\varphi_l^{style} = \varphi_l \varphi_l^T \qquad (5)$$

We then obtain the style loss by comparing $\varphi_l^{style}$ between $\varphi^{gt}$ and $\varphi^{out}$:

$$\mathcal{L}_{style} = \sum_{l=1}^{N} \frac{1}{C_l \times C_l} \left\| \frac{\varphi_l^{style_{gt}} - \varphi_l^{style_{out}}}{C_l H_l W_l} \right\|_1 \qquad (6)$$

*Identity Loss*: We enforce the identity similarity between the reconstructed face $\mathbf{I}_{out}$ and the original face $\mathbf{I}_{gt}$ in the embedding space which used to achieve **R5-R6**. The identity loss is formulated as follows

$$\mathcal{L}_{id} = \left\| E_{face}(\mathbf{I}_{gt}) - E_{face}(\mathbf{I}_{out}) \right\|_1 \qquad (7)$$

where $\| . \|_1$ is $\ell_1$-norm. The $E_{face}$ encoder is a pre-trained ArcFace model [31] with ResNet-50 backbone, trained on MS1MV2 dataset [44].

*Landmark Loss*: Because facial landmarks represent the potential poses of the face, we also include a sparse $L_2$ cycle consistency landmarks loss contributing to **R1-R4**. Using a pre-trained network named as $E_{lnd}$, landmarks are recovered. The landmark loss is formulated as follows

$$\mathcal{L}_{lnd} = \left\| E_{lnd}(\mathbf{I}_{gt}) - E_{lnd}(\mathbf{I}_{out}) \right\|_2 \qquad (8)$$

A pre-trained landmarks network ($E_{lnd}$) [50] has been used to predict 68 facial keypoints.

*Reconstruction Loss*: An additional loss is also used to encourage pixel-level reconstruction of $\mathbf{I}_{out}$. This loss is clearly motivated by our desire for $\mathbf{I}_{out}$ to be generally similar to $\mathbf{I}_{gt}$ and mainly address **R1-R4**. Notably, this loss can capture and preserve pixel-level information such as colors, illumination, and maintain texture information, not modeled by any other loss. It is calculated as the $\ell_1$-norm between $\mathbf{I}_{out}$ and the corresponding ground truth $\mathbf{I}_{gt}$. $\mathcal{L}_{rec}$ is defined as follows:

$$\mathcal{L}_{rec} = \alpha(1 - MS\_SSIM(\mathbf{I}_{gt}, \mathbf{I}_{out})) + (1 - \alpha)||\mathbf{I}_{gt} - \mathbf{I}_{out}||_1 \qquad (9)$$

where Multi-Scale Structural Similarity Index Metric (MS-SSIM) is calculated as in [13] and $\alpha = 0.84$.

*Adversarial Loss*: For adversarial loss, we use the non-saturating loss

with $R_1$ regularization [42]:

$$\mathcal{L}_{adv}^D = - \mathbb{E}_{w \sim \mathcal{W}}[\log D_w(w)] - \mathbb{E}_z[\log(1 - D_w(f(z)))] + \frac{\gamma}{2} \mathbb{E}_{w \sim \mathcal{W}}[\|\nabla_w D_w(w)\|_2^2] \qquad (10)$$

$$\mathcal{L}_{adv}^G = - \mathbb{E}_z[\log D_w(f(z))] \qquad (11)$$

*Total objective:* After defining the loss functions above, the total training objective can be expressed as:

$$\mathcal{L}_{total} = \lambda_{id}\mathcal{L}_{id} + \lambda_{lnd}\mathcal{L}_{lnd} + \lambda_{perc}\mathcal{L}_{perc} + \lambda_{style}\mathcal{L}_{style} + \lambda_{rec}\mathcal{L}_{rec} \qquad (12)$$

where $\lambda_{id}$, $\lambda_{lnd}$, $\lambda_{perc}$, $\lambda_{style}$ and $\lambda_{rec}$ are weights of corresponding losses, respectively. We set $\lambda_{id} = \lambda_{rec} = 1$, $\lambda_{lnd} = 0.001$, $\lambda_{style} = 0.1$ and $\lambda_{perc} = 0.01$ in our settings.

*Optimizer Loss:* Here we use ADAM optimization with Mean Square Error (MSE) and perceptual losses as the objective functions to find the optimal latent codes that can effectively approach $\mathbf{I}_{out}$ to $\mathbf{I}_{gt}$. This loss is aiding to address **R4-R6**. The loss function for optimization consists of two different loss terms including identity loss and perceptual loss:

$$\mathcal{L}_{opt} = \lambda_{perc}^o \mathcal{L}_{perc} + \lambda_{id}^o \mathcal{L}_{id} \tag{13}$$

We set $\lambda_{perc}^o = 0.01$ and $\lambda_{id}^o = 0.1$ in our settings.

Algorithm 1 shows the pseudo-code of the optimizer. Beginning with an appropriate initialization $w$, we look for an optimal vector $w^*$ that minimizes the $\mathcal{L}_{opt}$, which assesses how similar the given image and the image produced by $w^*$ are.

## 5. Experiments

The performance of the E2F-Net is assessed in this section using the newly created eyes-to-face datasets described in the next subsection. Our results have been compared with four methods: Pluralistic Image Completion (PIC) [26], EdgeConnect (EC) [17], LaFIn [6], and E2F-GAN [18]. To have fair comparison, the four methods have been retrained using the E2F-CelebA-HQ dataset. Five statistical metrics, described in subsection 4.3.1, have been used to quantitatively measure the performance difference among the methods. Additionally, we calculate the False Non-Match Rate (FNMR) between original and inpainted faces using a competitive face identification matcher [37] based on ArcFace [31] to assess the degree of retention of ID features.

### 5.1. Datasets

Experiments are conducted on seven generated datasets: E2F-StyleGANdb, E2F-CelebA-HQ, E2F-FFHQ, E2F-MS1MV2, E2F-LFW, E2F-CFP-FP, E2F-AgeDB-30; which are all available on the project's webpage. The images are resized to $256 \times 256$, and then a landmark detector [53] is used to locate and clip the eyes in order to extract the periocular area from each facial image. Furthermore, we eliminated deceptive samples utilizing WHENet methods [52], such as those with sunglasses over their eyes or faces that were tilted more than 45 degrees in one direction (roll, pitch, yaw), which would have hidden one of their eyes.

- *E2F-StyleGANdb:* A high-quality image dataset that consists of 50,000 pairs of $(\mathbf{I}_{gt}, \mathbf{I}_{in})$ images collected from StyleGAN outputs. We randomly selected 45,000 images for training and the remaining 5,000 images for testing. Each image has been resized to $256 \times 256$.

- *E2F-CelebA-HQ:* A high-quality image dataset that consists of 24,564 portrait images collected from a publicly available dataset [48]. We randomly selected 22,879 images for training and the remaining 1,685 images for testing. Each image has been resized to $256 \times 256$.

- *E2F-FFHQ:* A high-quality image dataset with more variations, consisting of 70,000 face images from a publicly available dataset known as FFHQ dataset [20]. All samples are used for testing the proposed E2F-Net. Each image has been resized to $256 \times 256$.

- *E2F-MS1MV2:* The original version of E2F-MS1MV2 called MS1MV2 [44] includes 85k identities and 5.8M images. After applying a landmark detector to extract the periocular region from each image, 83.8K identities and 5.6M images remained. This dataset has been used to train $E_{id}$.

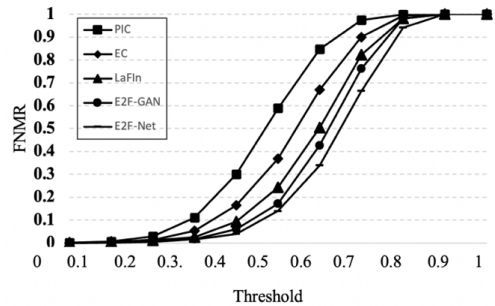

**Fig. 6.** FNMR curves are displayed for our proposed method (E2F-Net) and other compared methods (PIC, EC, LaFIn, E2F-GAN) using the E2F-CelebA-HQ dataset.

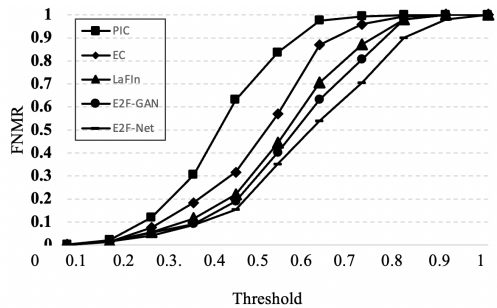

**Fig. 7.** FNMR curves are displayed for our proposed method (E2F-Net) and other compared methods (PIC, EC, LaFIn, E2F-GAN) using the E2F-FFHQ dataset.

- *E2F-LFW:* E2F-LFW is a modified version of LFW [45] including 6,000 pairs of faces in the validation part. After applying a landmark detector to extract the periocular region from each image, 5,996 pairs remained. This dataset has been used to evaluate $E_{id}$.

- *E2F-CFP-FP:* E2F-CFP-FP is a modified version of CFP-FP [46] including 6,000 pairs of faces in the validation part. After applying a landmark detector to extract the periocular region from each image, 5,998 pairs remained. This dataset has been used to evaluate $E_{id}$.

- *E2F-AgeDB-30:* E2F-AgeDB-30 is a modified version of AgeDB-30 [47] including 6,000 pairs of faces in the validation part. After applying a landmark detector to extract the periocular region from each image, 5,993 pairs remained. This dataset has been used to evaluate $E_{id}$.

## 5.2. Comparison Methods

In this work, we compare our method with four state-of-the-art inpainting methods, which are summarized as follows:

*PIC* [26]: PIC takes advantage of a dual pipeline using variational auto-encoders that consists of a reconstructive path that uses the ground truth to learn the prior distribution of missing regions and a generative path for which the conditional prior is connected to the distribution learned in the reconstructive path. It should be noted that, because the PIC approach produces distinct outputs for a certain input, it has been executed five times and the best outcomes have been reported.

*EC* [17]: By predicting the edges using an edge generator, EC feeds the damaged image and the predicted edges to the image inpainting network.

*LaFIn* [6]: LaFIn is an inpainting GAN-based network that uses predicted landmarks as guidance.

E2F-GAN [18]: E2F-GAN is a GAN-based coarse-to-fine method such that the coarse module benefits from the coarse-and-fine architecture. Moreover, an edge detector has been utilized to provide more information for the designed network.

## 5.3. Evaluation Metrics

Through quantitative and qualitative comparisons, we assess the proposed model's face inpainting performance. Two different types of metrics, comprising five statistics and one identity measure, have been calculated for quantitative analysis. We give a brief overview of each category and the associated metrics in the sections that follow.

### 5.3.1 Statistical Metrics

$\ell_1$ *loss* [20]. A simple and popular loss function used in the generation of images is the pixel-wise $\ell_1$ loss. This loss function measures the discrepancies between the synthesized content and the corresponding ground truth at the pixel level:

$$\ell_1(\mathbf{I}_{gt}, \mathbf{I}_{out}) = \frac{1}{hw} \sum_{i=1}^{h} \sum_{j=1}^{w} \left\| \mathbf{I}_{gt_{ij}} - \mathbf{I}_{out_{ij}} \right\|_1 \tag{14}$$

*Peak Signal to Noise Ratio (PSNR)* [38]:

$$PSNR\,(\mathbf{I}_{gt}, \mathbf{I}_{out}) = 10\,log_{10}^{\frac{255^2}{\frac{1}{hw}\sum_{i=1}^{h}\sum_{j=1}^{w}(I_{gt_{ij}} - I_{out_{ij}})^2}} \tag{15}$$

*Structural Similarity (SSIM)* [13]. The SSIM describes the degree of structural similarity between two images:

$$SSIM(\mathbf{I}_{gt}, \mathbf{I}_{out}) = \frac{(2\mu_{gt}\mu_{out} + C_1)(2\sigma_{gt}\sigma_{out} + C_2)}{(\mu_{gt}^2 + \mu_{out}^2 + C_1)(\sigma_{gt}^2 + \sigma_{out}^2 + C_2)} \tag{16}$$

where $C_1$ and $C_2$ are positive constants added to prevent cases in which the denominator is zero.

*Frechet Inception Distance (FID)* [30]. Utilizing the Wasserstein distance between the distributions of the actual and created images in the feature space acquired by the Inception model [30], this metric assesses the visual quality and variety of the generated images. The FID can be expressed as:

$$FID\,(\mathbf{I}_{gt}, \mathbf{I}_{out}) = \left\| \mu_{gt} - \mu_{out} \right\|_2^2 + Tr(\sigma_{gt} + \sigma_{out} - 2\left(\sigma_{gt}\sigma_{out}\right)^{\frac{1}{2}}) \tag{17}$$

In both SSIM and FID metrics, $\mu_{gt}$ and $\mu_{out}$ donate the mean values of $\mathbf{I}_{gt}$ and $\mathbf{I}_{out}$, respectively; while $\sigma_{gt}$ and $\sigma_{out}$ represent the covariance of $\mathbf{I}_{gt}$ and $\mathbf{I}_{out}$, respectively.

*Total Variation (TV) [37]*. TV calculates the total of the absolute differences for nearby pixels as formulated below to help quantify the degree of noise in the image:

$$TV(\mathbf{I}_{gt}, \mathbf{I}_{out}) = \sum_{i,j}^{h-1,w} \frac{\left\| \mathbf{I}_{out_{i+1,j}} - \mathbf{I}_{out_{i,j}} \right\|_1}{N_h} + \sum_{i,j}^{h,w-1} \frac{\left\| \mathbf{I}_{out_{i,j+1}} - \mathbf{I}_{out_{i,j}} \right\|_1}{N_w}, \tag{18}$$

where $N_h$ and $N_w$ are the number of pixels in $\mathbf{I}_{out}$ except for the last row and the last column, respectively.

### 5.3.2    Identity Metrics

We used false non-match rate (FNMR) to measure the preserved ID attributes in inpainted images. More specifically, FNMR measures the miss-categorization rate for some pairs of face images where each pair belongs to the same individual. Here, we assumed that $\mathbf{I}_{in}$ and $\mathbf{I}_{out}$ are two faces for the same individual. Using $E_{face}$, the corresponding embedding vectors for each face have been obtained, and the cosine similarity for each pair of $\mathbf{I}_{in}$ and $\mathbf{I}_{out}$ has been calculated. Finally, the FNMR for different thresholds has been depicted.

### 5.4. Implementation Details

We use StyleGAN pre-trained at 256x256 resolution in all our experiments. Training is performed using the Adam optimizer, with $\beta_1 = 0.9$ and $\beta_2 = 0.999$. On a single NVIDIA GeForce RTX 3090 GPU, the network is trained end-to-end with batch sizes of 16 and converges in roughly two days. It should be noted that this is quite effective considering that training StyleGAN would take more than 35 days on the same GPU.

### 5.5. Comparison with Previous Works

We qualitatively and quantitatively compare our results against four state-of-the-art approaches, i.e., PIC, EC, LaFIn, and E2F-GAN, using the above-mentioned metrics and plotting some outputs. It should be noted that we trained the four methods using our own constructed training dataset, E2F-CelebA-HQ, using the best reported setups for each method described in the respective articles. The obtained results based on the E2F-CelebA-HQ and E2F-FFHQ validation datasets have been presented in the following subsections.

**Table 2.** Quantitative results over E2F-CelebA-HQ dataset for E2F-Net and other compared methods (PIC, EC, LaFIn, E2F-GAN). The best result of each column is boldfaced. ↑ indicates that the higher the number the better is the model and ↓ indicates the lower the number the better is the model.

| Method | FID ↓ | SSIM ↑ | PSNR ↑ | TV ↓ | $\ell_1 Loss$ ↓ |
|---|---|---|---|---|---|
| PIC | 57.02 | 0.41 | 11.19 | 8.50 | 50.37 |
| EC | 70.63 | 0.42 | 12.67 | 5.27 | 121.08 |
| LaFIn | 63.16 | 0.47 | 13.18 | 6.89 | 40.94 |
| E2F-GAN | 46.39 | 0.51 | 13.66 | **0.02** | 41.54 |
| **E2F-Net (Ours)** | **45.85** | **0.53** | **13.78** | **0.02** | **40.36** |

**Table 3.** Quantitative results over E2F-FFHQ dataset for E2F-Net and other compared methods (PIC, EC, LaFIn, E2F-GAN). The best result of each column is boldfaced. ↑ indicates that the higher the number the better is the model and ↓ indicates the lower the number the better is the model.

| Method | $FID \downarrow$ | $SSIM \uparrow$ | $PSNR \uparrow$ | $TV \downarrow$ | $\ell_1\ Loss \downarrow$ |
|---|---|---|---|---|---|
| PIC | 143.89 | 0.37 | 10.03 | 10.54 | 68.9 |
| EC | 134.63 | 0.37 | 10.84 | 7.59 | 197.84 |
| LaFIn | 97.48 | 0.43 | 11.32 | 6.98 | 55.29 |
| E2F-GAN | 101.27 | 0.45 | 11.52 | **0.02** | 53.64 |
| **E2F-Net (Ours)** | **91.14** | **0.49** | **12.12** | **0.02** | **49.12** |

### 5.5.1. Quantitative Comparisons

The results of statistical metrics calculated on the validation set of the E2F-CelebA-HQ dataset including 1,685 samples are reported in Table 2. As can be observed, E2F-Net is superior over PIC, EC, LaFIn, and E2F-GAN in most metrics except TV loss which is equal to E2F-GAN. Overall, the E2F-Net outperforms the other methods in terms of FID, SSIM, PSNR, and $\ell_1$ metrics (addressing **R1-R4**). More precisely, the significant margins in FID and $\ell_1$ measures show that, in comparison to previous approaches, our method can inpaint the masked image with a significantly greater level of quality. The large margin between our proposed method and others is also patent in Table 3, when the E2F-FFHQ dataset has been used as a validation set. We conducted t-tests to statistically validate the E2F-Net model's superior performance in metrics such as FID, SSIM, PSNR, TV, and $\ell_1$ loss, against other methods with p-values under 0.05. Our statistical analysis revealed that our method outperforms other techniques across various metrics for both datasets, with the exception of the TV metric. Specifically, the TV metric showed no significant statistical difference when comparing our method with the E2F-GAN across both E2F-CelebA-HQ and E2F-FFHQ datasets.

Moreover, to measure the amount of preserving ID features, FNMR has been calculated for both datasets as shown in Figs. 6 and 7 (addressing **R6**). E2F-Net has a decreased false non-match rate at various thresholds, demonstrating the capability of our system to extract ID from the periocular area and transfer it to the reconstructed face.

Additionally, to quantitatively measure the preserved demographic information (e.g., age, gender) (addressing **R5**), we used OpenCV age and gender estimation library, to compare the reconstructed faces with ground truth. Notably, the E2F-CelebA-HQ, the modified version of CelebA-HQ, is an ill-biased/imbalanced dataset over certain attributes such as gender, skin color, and age.

Regarding gender, 66 percent of training images are female, and 34 percent are male, indicating female gender representation over male gender during training. Furthermore, this imbalance also exists for validation images, 58 percent are female and 42 percent are male. Regarding age, four intervals are considered, and the percentage of each interval is reported in Table 4. The training and validation sets are imbalanced such that the second interval (i.e., between 15 and 40) has the maximum samples ($\approx$ 55%) and the fourth interval (i.e., older than 60) has the minimum samples (i.e., $\approx$ 0.1% in training set and $\approx$ 1% in validation set). Considering the validation set percentage as the baseline for each interval, our proposed method (E2F-Net) predicts the age attribute very close to the baseline. Regarding gender, as shown in Table 5 our proposed method along with E2F-GAN and LaFIn methods can preserve the attributes of both groups very well (i.e., 41.7% out of 42% of men and 57.9 % out of 58% of women).

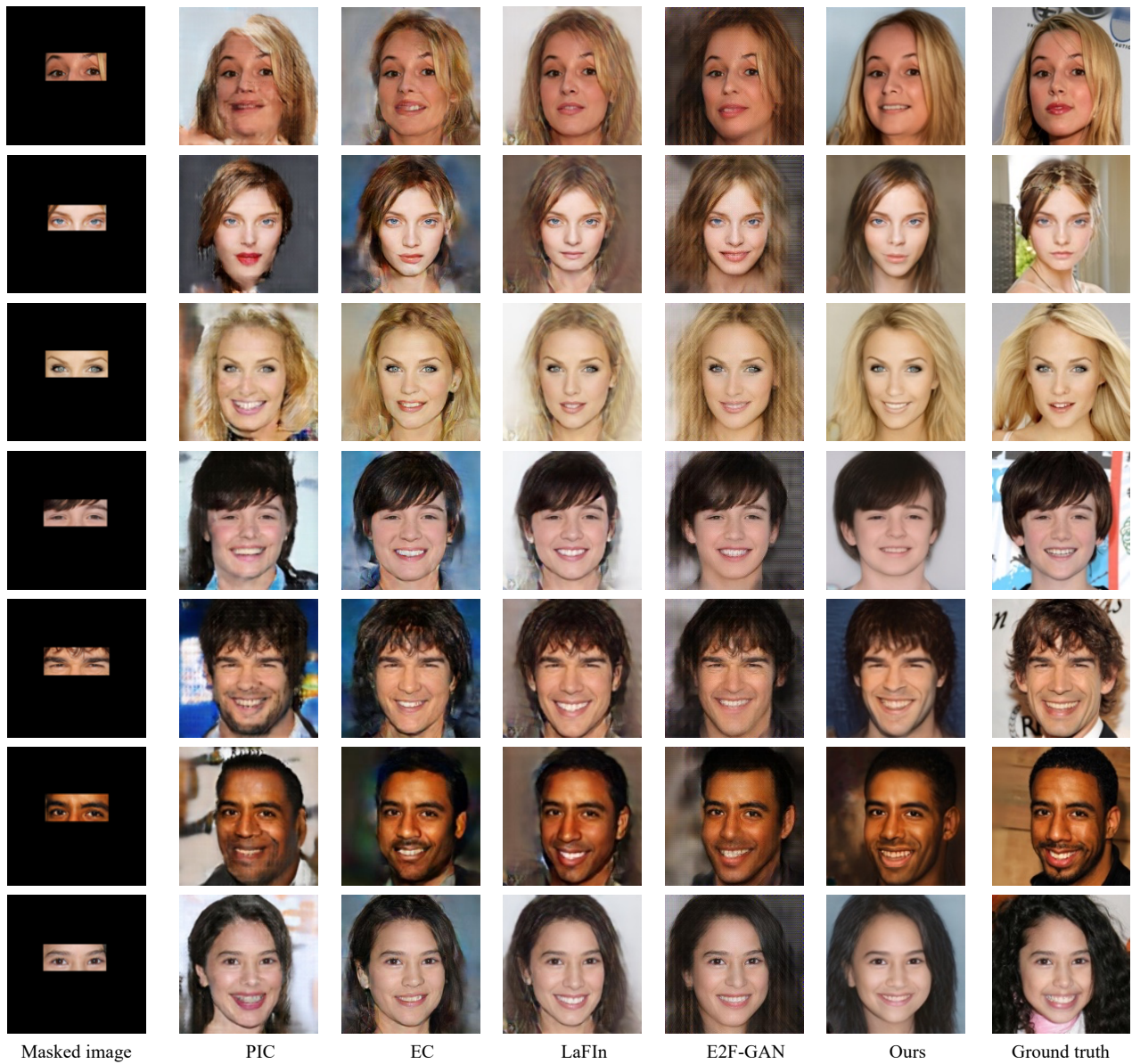

| Masked image | PIC | EC | LaFIn | E2F-GAN | Ours | Ground truth |

**Fig. 8.** Quality comparison among PIC, EC, LaFIn, E2F-GAN, and our proposed method using E2F-CelebA-HQ dataset.

### 5.5.2. Qualitative Comparisons

A few samples of results are displayed in Figs. 9 and 10 for E2F-CelebA-HQ and E2F-FFHQ datasets, respectively. As it can be observed, the quality of PIC and EC is really low compared to E2F-GAN, LaFIn, and our results. Moreover, in comparison with other methods, our method generates high quality and highly structured faces (addressing **R1-R4**). Additionally, to measure the amount of preserved demographic information (addressing **R5**), we present a variety of faces in Figs. 8 and 9. For instance, Fig. 8 rows 2, 4, and 7 show very young man and women with faces reconstructed preserved well ID features. Similar high-quality results are demonstrated in Fig. 10 rows 5 and 6 for two elderly men.

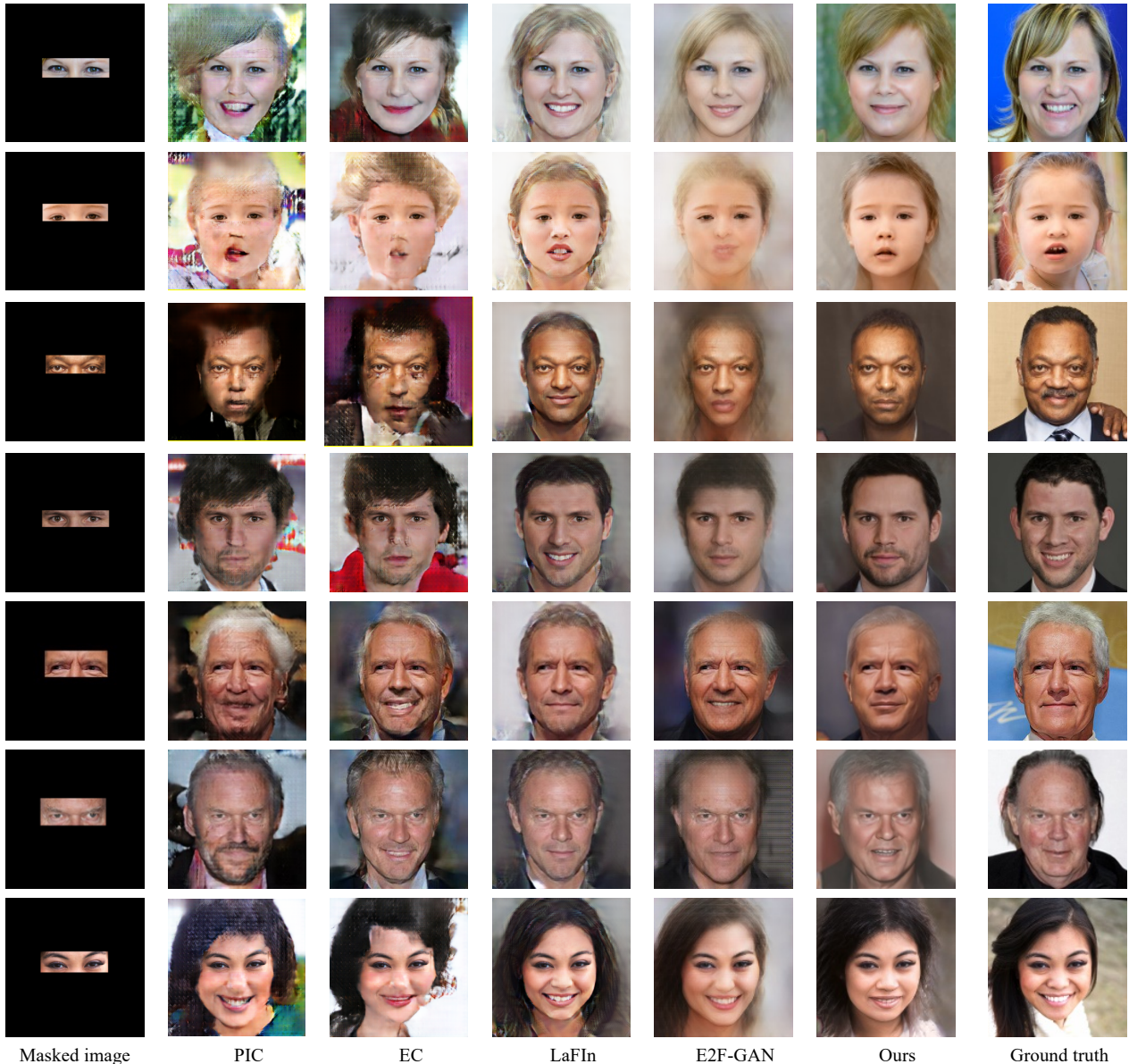

| Masked image | PIC | EC | LaFIn | E2F-GAN | Ours | Ground truth |

**Fig. 9.** Quality comparison among PIC, EC, LaFIn, E2F-GAN, and our proposed method using E2F-FFHQ dataset.

## 6. Ablation Study

*Number of layers of the mapper ($M$).* A fully-connected network has been used to map $\mathcal{Z}$ space latent codes to $\mathcal{W}$ space. Notably, in the original StyleGAN network an eight-layer fully connected network was proposed. To analyze the optimum mapper for this task, we have done the experiments using a different number of layers: 2, 4, and 8 layers. As shown in Table 6, a 4-layer fully-connected mapper generates better quantitative results.

**Table 4.** Age Evaluation of our proposed method (E2F-Net), PIC, EC, LaFIn, and E2F-GAN using CelebA-HQ dataset. numbers are shown in percent.

|  | Age <= 15 | 15 < Age <= 40 | 40 < Age <= 60 | Age > 60 |
|---|---|---|---|---|
| E2F-CelebA-HQ (training set) | 38.1 | 54.8 | 7 | 0.1 |
| E2F-CelebA-HQ validation set (baseline) | *37* | *55* | *7* | *1* |
| PIC | 24 | 48.9 | 5.6 | 0.4 |
| EC | 16.4 | 45.6 | 4.4 | 0.2 |
| LaFIn | 34.8 | 53.2 | 6.3 | 0.7 |
| E2F-GAN | 34.8 | 53.2 | 6.2 | 0.8 |
| **E2F-Net (ours)** | **35.1** | **53.9** | **6.5** | **0.8** |

**Table 5.** Gender Evaluation of our proposed method (E2F-Net), PIC, LaFIn, EC, and E2F-GAN using CelebA-HQ dataset. numbers are shown in percent.

| | *male* | *female* |
|---|---|---|
| CelebA-HQ training set | *34* | *66* |
| CelebA-HQ validation set (baseline) | *42* | *58* |
| PIC | 40.6 | 56.8 |
| EC | 40.1 | 56.6 |
| LaFIn | 41.7 | 57.9 |
| E2F-GAN | 41.7 | 57.9 |
| **E2F-Net (ours)** | **41.7** | **57.9** |

*Impact of optimizer:* Table 7 shows quantitative metrics for the initial output of StyleGAN and after executing our latent embedding optimizer. We show the results of $25^{th}$, $100^{th}$, and $200^{th}$ iterations. Utilizing an RTX 3090 graphics card, we recorded the optimization time consumed after the $25^{th}$, $100^{th}$, and $200^{th}$ epochs, detailed in Table 7 and discussed in [28]. Notably, the consumed time before initiating the optimization process is 0.48 second.

For all metrics, the optimizer has a positive impact. Fig. 10 shows the impact of the optimizer on improving quality, identity features, and facial expression.

*Impact of our inpainting method on face/periocular recognition*: To measure if our proposed inpainting method preserves or adds further ID information in the reconstructed face, we created two datasets and used $E_{id}$ (i.e., trained on periocular region) and $E_{face}$ (i.e., trained on face) to calculate FMR and FNMR curves.

**Table 6.** Quantitative results over E2F-CelebA-HQ dataset using 2, 4, 8-layer mapper. for each metric, a triplet including the initial output of StyleGAN, the output of StyleGAN after 25 and 200 iterations on $f$ have been reported respectively.

| Method | *FID ↓* | *SSIM ↑* | *PSNR ↑* | *TV ↓* | *$\ell_1$Loss ↓* |
|---|---|---|---|---|---|
| *2-layer mapper (initial/25/200)* | 54.19/50.87/48.61 | 0.47/0.49/0.49 | 13.62/13.57/13.38 | 0.02/0.018/0.019 | 40.77/39.77/40.25 |
| *4-layer mapper (initial/25/200)* | **47.85/46.37/45.85** | **0.47/0.49/0.49** | **13.67/13.66/13.44** | **0.02/0.018/0.019** | **40.36/39.41/40.05** |
| *8-layer mapper (initial/25/200)* | 50.02/48.10/46.07 | 0.47/0.49/0.49 | 13.49/13.50/13.27 | 0.02/0.018/0.019 | 41.38/40.03/40.67 |

**Table 7.** the quantitative results of the initial outputs of StyleGAN and after optimizing the latent code in three different iterations over E2F-CelebA-HQ dataset and consumed time using 4-layer mapper have been reported.

| Method | *FID ↓* | *SSIM ↑* | *PSNR ↑* | *TV ↓* | *$\ell_1$Loss ↓* | Time [sec] |
|---|---|---|---|---|---|---|
| Initial output | 49.8586 | 0.479774 | 13.6780 | 0.0208 | 40.360 | 0.48 |
| $25^{th}$ iteration | 48.3770 | 0.49789345 | 13.6619 | 0.0186 | 39.417 | 2.04 |
| **$100^{th}$ iteration** | **47.1548** | **0.5002335** | **13.5310** | **0.0181** | **39.384** | 6.4 |
| $200^{th}$ iteration | 47.2831 | 0.49711797 | 13.4402 | 0.0197 | 40.056 | 12.7 |

We used CelebA-HQ validation set to create the required datasets. First, a dataset called sub-CelebA including 960 face pairs (480 positive pairs such that both images are belonging to the same subject and 480 negative pairs in which the images of each pair are belonging to different subjects) have been selected. Then all images within the sub-CelebA dataset are cropped such that just the periocular region remained, we called this dataset p-CelebA (p for periocular). Finally, again using sub-CelebA, we create the inpaint-CelebA dataset such that one of the faces in each pair is kept, and the other one is replaced with its reconstructed version.

Therefore, to measure the performance of E2F-Net regarding preserving ID information, we feed the p-CelebA to $E_{id}$ and inpaint-CelebA to the $E_{face}$. The results including the angle distributions, the FMR, FNMR, and ROC curves for the above-mentioned datasets are shown in Fig. 11. The FNMR curves drawn using the $E_{face}$ increase more smoothly compared to FNMR curves drawn by $E_{id}$, and the FMR curve drawn by $E_{face}$ decreases more quickly compared to FMR curves drawn by $E_{id}$, leading to less EER value (crossing point between FMR and FNMR curves). Notably, the large gap between FMRs curves demonstrates that the inpainting-based face recognition can reduce inter-class distances significantly. We hypothesize it as a result of our proposed architecture being able to extract a great amount of ID information from the periocular region and transmit to inpainted face. The accuracies for p-CelebA and inpaint-CelebA datasets are 90.81% and 96.04%, respectively.

*Impact of other types of masks:* while the current study focused on eyes-to-face task, we have checked the capability of E2F-Net for four other types of masks including free-form and fixed-form. Notably, since $E_{id}$ is trained to extract identity information from periocular region, we cannot use masks that cover this region. The results are presented in Fig. 12.

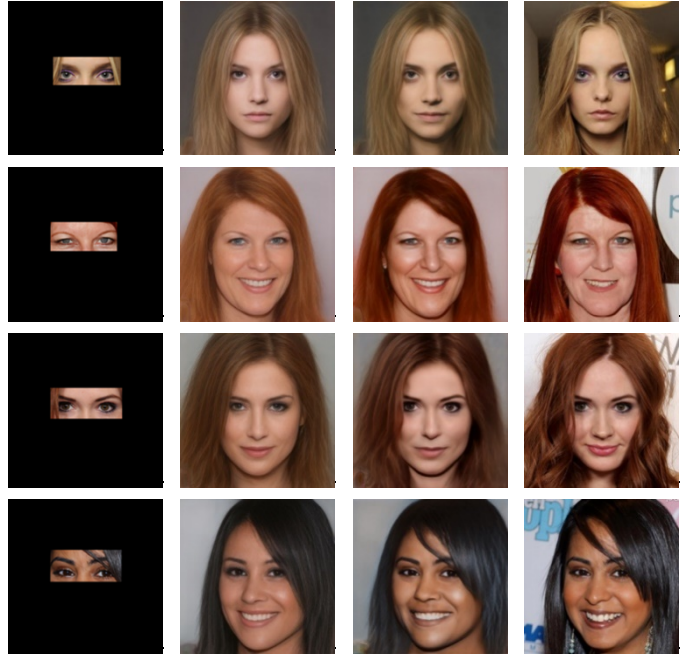

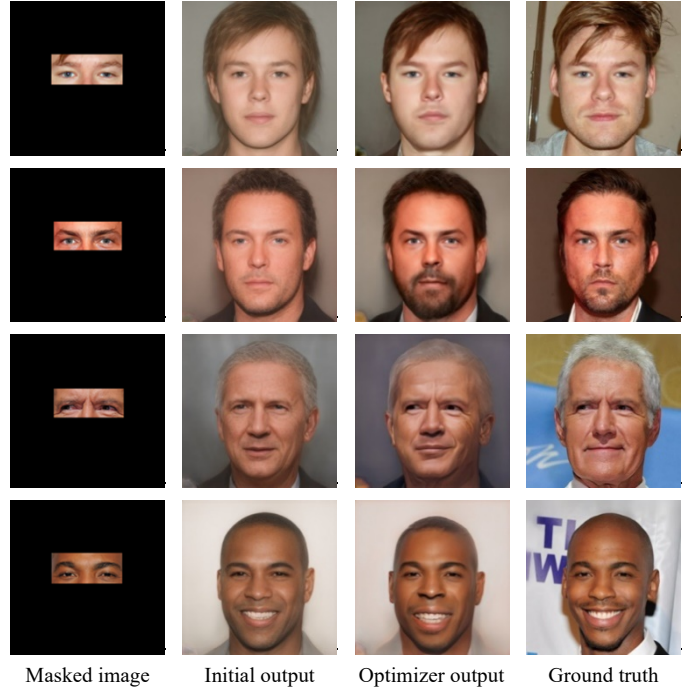

Masked image    Initial output    Optimizer output    Ground truth

**Fig. 10.** Impact of optimizer on improving quality, ID features, and facial expression of StyleGAN output.

## 7. Discussion

Regarding preserving the ID information of each person (i.e., addressing **R6**), we have done two main experiments. First, the outputs of our proposed method are compared with other methods by calculating FNMR curves on two datasets (i.e., E2F-CelebA-HQ and E2F-FFHQ) shown in Figures 7 and 8. Second, we have explored if inpainted full faces are more adequate for person ID compared to the input periocular images. Our conducted experiments (i.e., in section V) show that inpainted full face recognition improves the verification performance over periocular-based person verification. Common sense tells us that 1) inpainting properly keeps ID information, and 2) the state-of-the-art face recognition models work better in full face compared to periocular images.

*Limitations of our Work*: Although our proposed method can reconstruct a wide variety of faces while preserving ID features, there are two main limitations. First, the color of the scalp hair and eyebrow hair can result in different colors (as shown in Fig. 13 row 1). Detecting and properly inpainting these elements may not be feasible without other cues if, e.g., part of the scalp hair is not visible in the periocular region, if the person hides part of the scalp hair (e.g., by a hat as shown in Fig. 13 row 2), or misses part the of scalp hair (see Fig. 13 row 3). Another issue for men is the difficulty to detect the existence of a beard on the face based on the periocular region. Second, the existence of occlusion or closed eyes may lead to reducing the quality of outputs as shown in Fig. 14.

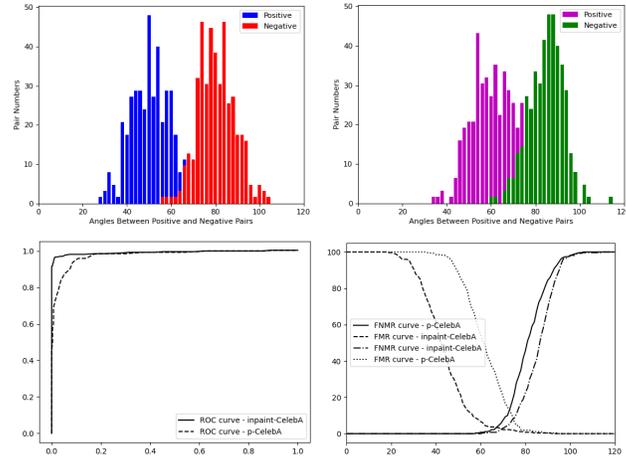

**Fig. 11.** Comparing the performance of $E_{face}$ and $E_{id}$ fed by inpaint-CelebA and p-CelebA datasets, respectively. The first row shows the angle distributions for positive and negative pairs for both inpaint-CelebA (left) and p-CelebA (right) datasets. The ROC, FMR and FNMR curves for both datasets are shown in the second row.

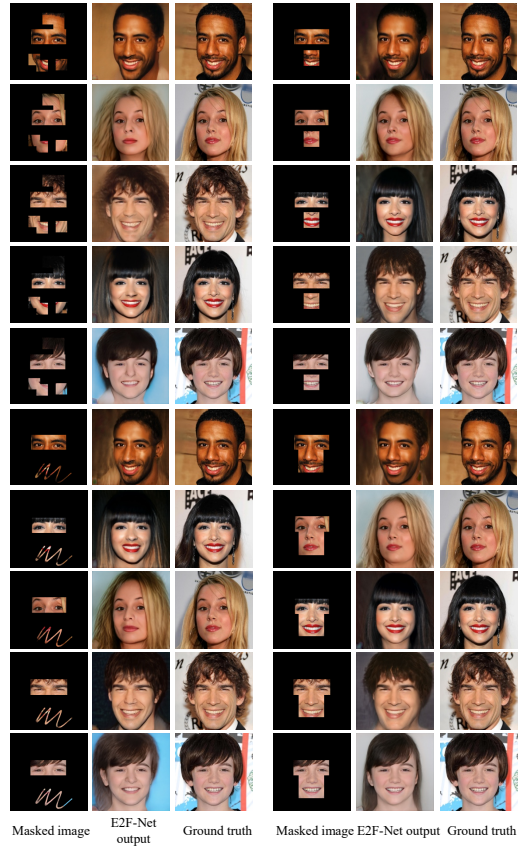

| Masked image | E2F-Net output | Ground truth | Masked image | E2F-Net output | Ground truth |

**Fig.12.** Comparative analysis of E2F-Net's performance with various mask types.

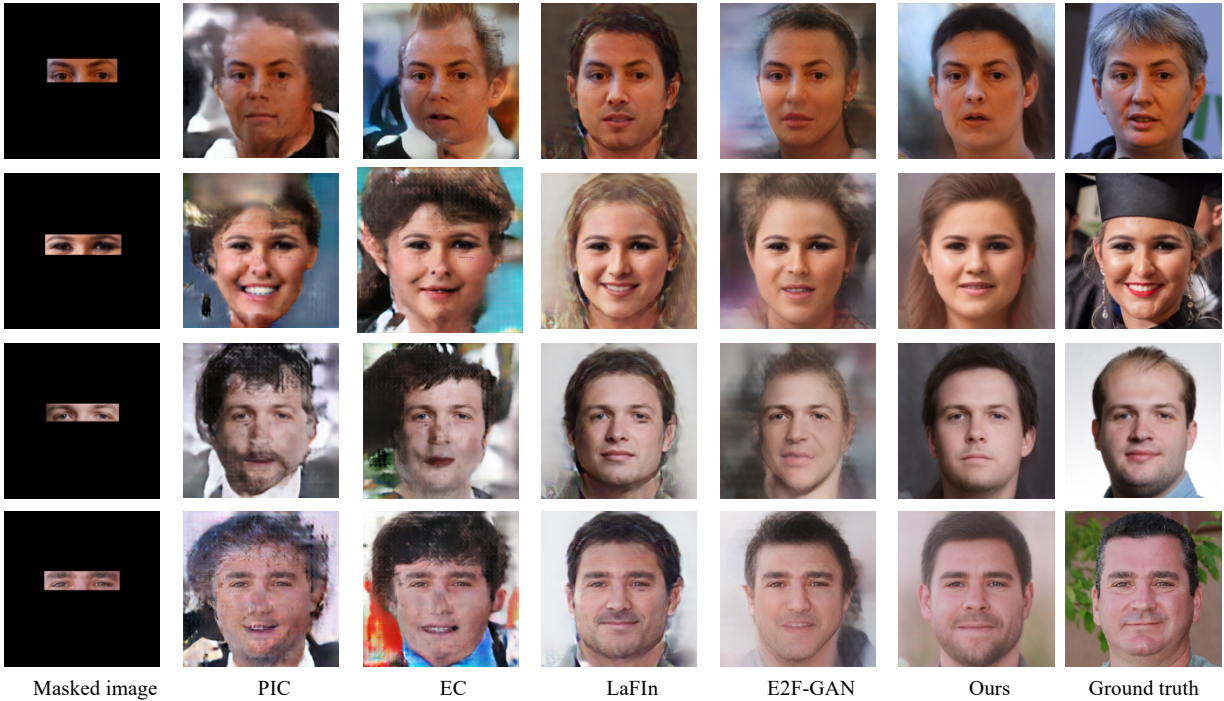

| Masked image | PIC | EC | LaFIn | E2F-GAN | Ours | Ground truth |

**Fig. 13.** Quality comparison among PIC, EC, LaFIn, E2F-GAN, and our proposed method using E2F-FFHQ dataset.

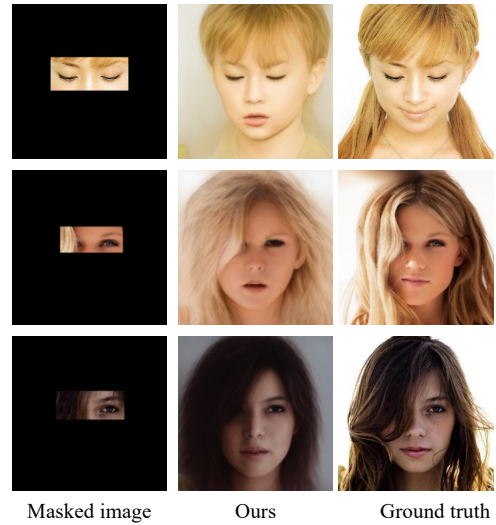

| Masked image | Ours | Ground truth |

**Fig. 14.** Example impact of closed eyes and eyes occlusion on output.

## 8. Conclusion and Future Work

In this paper, we show that a variety of faces can be reconstructed using only the periocular region by our proposed

GAN-based network called E2F-Net. For this purpose, a pre-trained face generator called StyleGAN has been used such that our proposed method benefits from not only minimum training process but also high image quality and diverse facial outputs. Moreover, to carefully extract ID features from the periocular region, we used a face recognition model called ArcFace which is retrained on

E2F-MS1MV2 dataset, a generated identity recognition dataset based on the eyes region. Notably, we reveal that ID and non-ID

features can be extracted from the eyes region and finally reconstruct the whole face based on these features. We conducted extensive experiments on two datasets including a high diversity of faces with different gender, ethnicity (e.g., Caucasian, Asian, African), pose (e.g., frontal, upward, downward), and expression (e.g., smiling, neutral), and show that our method successfully reconstructs the whole face with high quality.

Despite promising effectiveness, the proposed method still needs to be further improved: (1) the capacity of the generated GAN latent space through adversarial loss to represent the space from ground-truth data is challenging to measure. While current experiments with different mapper and discriminator architectures provide some insights, there's still uncertainty about the adequacy of the latent space representation. Future work should explore novel ways to validate the representation capacity of the GAN latent space. (2) The size of the training dataset and its influence on generalization is another concern. With the current architecture relying on only three trainable components and relatively shallow networks for $M$, and $D_w$, the impact of a larger dataset may be limited. However, future studies could explore scaling up the architecture to leverage larger datasets more effectively. (3) To address the challenges associated with hair and eyebrow color consistency, the presence of occlusions, and the detection of facial hair from limited visual information. These factors currently impede the method's reliability in reconstructing facial features with high fidelity. (4) by systematically manipulating various facial elements, future research could yield valuable insights into the differential contributions of ID and non-ID features to facial recognition. (5) we fine-tuned the model's trainable components using high-resolution facial datasets (i.e., E2F-CelebA-HQ) which resulted in the model being specialized for high-resolution imagery. However, certain use cases, like public security and criminal identification, may not always have high-resolution images available. Therefore, subsequent research could focus on refining our model to perform effectively in situations where only lower-resolution images are accessible.